# $2^B 3^C$: 2 Box 3 Crop of Facial Image for Gender Classification with Convolutional Networks

Vandit Gajjar

Department of Electronics and Communication Engineering and
Computer Vision Group, L. D. College of Engineering, Ahmedabad, India

`gajjar.vandit.381@ldce.ac.in`

## Abstract

*In this paper, we tackle the classification of gender in facial images with deep learning. Our convolutional neural networks (CNN) use the VGG-16 architecture [1] and are pretrained on ImageNet for image classification. Our proposed method (2B3C) first detects the face in the facial image, increases the margin of a detected face by 50%, cropping the face with two boxes three crop schemes (Left, Middle, and Right crop) and extracts the CNN predictions on the cropped schemes. The CNN of our method is fine-tuned on the Adience and LFW with gender annotations. We show the effectiveness of our method by achieving 90.8% classification on Adience and achieving competitive 95.3% classification accuracy on LFW dataset. In addition, to check the true ability of our method, our gender classification system has a frame rate of 7-10 fps (frames per seconds) on a GPU considering real-time scenarios.*

## 1. Introduction

Gender classification based on facial images has always been one of the most studied soft-biometric subjects. Automatic face detection and classification of gender using machine learning models [2, 3, 4, 5, 6, 7, 8, 9] has gained high attention for more than two decades. In fact, many researchers still use these models in the less constrained environment, to improve gender classification error-rate. Since the Deep Convolutional Neural Networks (D-CNN) such as AlexNet [10] introduced by Krizhevsky *et al.*, this domain has largely replaced the need for handcrafted facial descriptors. D-CNN models not only successfully applied for face and human analysis, but also for video classification (Karpathy *et al.*) [11], audio classification (Lee *et al.*) [12], text classification (Lai *et al.*) [13] and many more classification tasks.

In previous five years, considering gender classification on constrained faces, the classification accuracies increased gradually for different datasets such as 91% (Dehghan *et al.*) [14] On Adience [15], 95.98% (Afifi *et al.*) [16] On LFW [17], 99.90% (Moeini *et al.*) [18] On FERET [19]. However, on less constrained faces such as an occluded face, low-resolution facial image and in changed viewpoint, gender classification is still facing several challenges. Sample images are shown in Fig. 1.

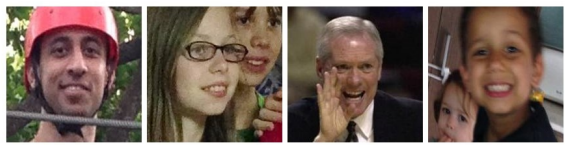

Figure 1: Sample images having less constrained environments i.e. occlusion, low resolution.

Our CNN use the VGG-16 architecture [1] and is pretrained on ImageNet [21] for image classification task. Thus, we gain the advantage from the representation learned to differentiate object types from images. Fine-tuning the CNN on training images with gender annotations is an essential step to gain advantage from the representation power of the CNN.

In this work, we tackle the gender classification problem from different angle. We are inspired by the recent advances in the area of image classification using D-CNN to achieve state-of-the-art results. Based on the above motivation our work for gender classification is as follows:

Our $2^B 3^C$ method, first detect the face in the facial image using robust face-detector proposed by Li *et al.* [20], increases the detected margin by 50%, face cropping with three crop schemes (Left, Middle, and Right crop), passed into CNN and then extract the CNN predictions by summing three crop result. Our method is fine-tuned on our prepared[1] Adience and LFW dataset.

Our main contributions of this paper are as follows:

- Using robust face-detector, we propose CNNs based method using two box three crop schemes for facial gender classification. It achieves the gender classification accuracy with 90.8% and 95.3% on Adience and LFW dataset respectively.

---

[1] Section 3.3 describes the preparation of train, validation and test image-set.

- Using GPU support, we test the true ability of our method by gaining 7-10 fps (frames per seconds) on real-time scenarios for gender classification task.

The rest of this paper is as follows. In Section 2 we discuss the related work about face-detection, D-CNN and gender classification. In Section 3 we describe the proposed method with pre-processing, preparation of training and testing data and implementation details. The experimental results are briefly shown in Section 4. Section 5 concludes the paper.

## 2. Related Work

In this section, we briefly review related work for D-CNN, Face Detection and Gender Classification.

### 2.1. Deep Convolutional Neural Networks

One of the first applications of CNN is LeNet [22] proposed by Lecun et al. for handwritten code recognition. Their model was relatively simple due to the limited computational power and the algorithmic challenges of training bigger networks. Since the computational power and ImageNet database [23] introduced by Deng et al. the growth of D-CNN has increased exponentially. After the Large-scale Visual Recognition Challenge [21] has been proposed several deeper architectures won the challenge i.e. AlexNet [10], VGG-16 [1], GoogLeNet [24], ResNet [25], DenseNet [26], AOGNet [27].

D-CNN models have been shown to perform well despite the strong difference in changing viewpoints, and lighting conditions. The strength of CNN and their success in the task of Face-detection make them a reasonable candidate for facial gender classification system.

### 2.2. Face Detection

Remarkable performance of D-CNN on image classification [21, 28, 29] and object detection [30, 31, 32, 33], current face-detection studies [20, 34, 35, 36, 37, 38] also embrace deep learning for better performance. These proposed methods use D-CNN as the backbone structure to learn highly differentiable representation from data and achieve competitive results on benchmark datasets such as FDDB, LFW, AFW, WIDER Face, IMDB-Wiki. Among these methods, Faceness-Net [37], Faster R-CNN [38], and Multi-View detection [35] are created to detect faces under heavy occlusions, changing viewpoints, different lighting condition and large pose variations. Cascade-CNN proposed by Li et al. [20] achieves an excellent trade-off between speed and accuracy. In our work, we have used the Cascade-CNN proposed by Li et al. [20] as the detector is several times faster than the model-based and exemplar-based detection systems and has a frame-rate of 14 fps on CPU and achieves 100 fps using GPU.

### 2.3. Gender Classification

A detailed survey of the gender classification methods can be found in [39]. Here we quickly discuss the appropriate methods/approaches.

Gender is one of the key facial attribute, which plays a fundamental role in our social interactions. One of the approaches using neural networks proposed by Golomb et al. [40] for gender classification, which was trained on a small set (90 images) of facial images. Since then numerous approaches have been proposed using support vector machines, and D-CNNs.

For unconstrained environments, Nilan et al. [41] proposed robust gender classification system using CNNs. They have achieved the competitive result on LFWA [17] with the classification accuracy of 98.8%. In comparison with large-scale datasets for object detection and segmentation, like ImageNet [23], MS-COCO [28], PASCAL VOC [29], there are limited small-scale datasets are available for gender classification such like Adience [15], LFW [17], IMDB-WIKI [42]. Levi et al. propose a CNN based approach that works well on this type of small dataset.

Adience is one of the popular datasets for facial gender classification and age estimation, which was published by Eidinger et al. [15] in 2014, containing 26580 images of 2284 subjects with gender and age annotations. Same as LFW and IMDB-Wiki dataset is also recognizable for face detection and gender classification tasks. Still, the reason behind the Adience to be used in many testing protocols is the dataset is containing images, which are very closely related to real-world scenarios. The dataset includes several variations like the change in appearance, pose, different light exposure, low and high-quality images. These types of conditions provide a less constrained environment and challenging learning problem for algorithms/methods. In competition with Adience, LFW is also a key dataset for gender classification having 13233 images of 5749 subjects (1680 subject having two or more images). The dataset also contains same less constrained quality as given in Adience. In our work, we have selected these two datasets, as these two datasets are closely related to real-time scenarios.

For the first time, a D-CNN model was applied to Adience by Levi et al. [43]. Using simple network architecture, they have achieved competitive result with 86.8% classification accuracy. As the Adience and LFW is challenging datasets for evaluation, Liao et al. [44] has proposed nine-patch method which has gain 78.63% and 95.64% classification accuracy on Adience and LFW respectively. To the best of our knowledge, the current state-of-the-art result on Adience and LFW datasets are reported by Dehghan *et al.* [45] with 91% and Jia *et al.* [47] with 98.9% for facial gender classification. Table 1; shows an overview of the development of gender classification on Adience and LFW datasets.

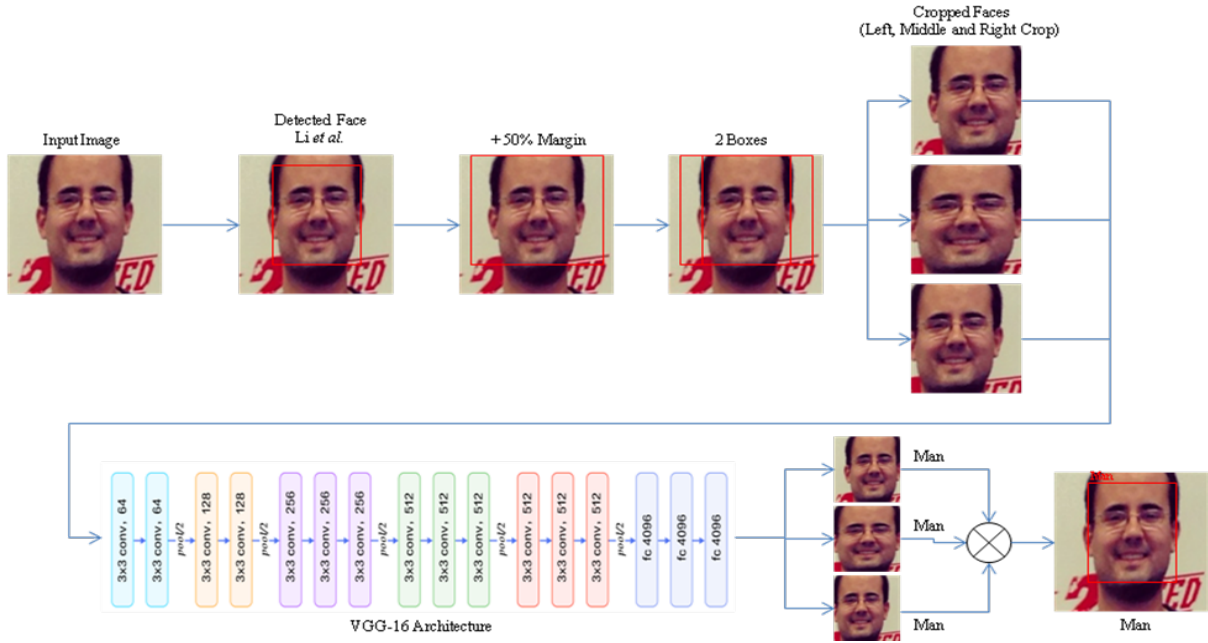

Figure 2: The pipeline of our method for gender classification.

| Adience | | LFW | |
|---|---|---|---|
| Approach | %Accuracy | Approach | %Accuracy |
| SVM [15] | 77.8 | LBP [49] | 90 |
| SVM [44] | 78.63 | SVM [48] | 94.81 |
| SVM [46] | 79.3 | SVM [44] | 95.64 |
| D-CNN [43] | 86.8 | D-CNN [41] | 98.8 |
| D-CNN [45] | 91.0 | D-CNN [47] | 98.9 |

Table 1: Gender classification accuracy, using different approaches on Adience and LFW dataset.

## 3. Our Proposed Method ($2^B3^C$)

The detailed pipeline of our method is shown in Figure 2. It includes four stages as follows:

➢ Firstly, in the given facial image, we detect the face using robust face-detector proposed by Li *et al.* [20].

➢ Then with the detected face, we increase the bounding box by 50% of margin. (50% of its height above and 50% of its width right and left.)

➢ Then on the increased marginal detected face, we create two boxes and crop three faces, scaled to 224x224 resolutions.

➢ The cropped faces are fed to VGG-16 architecture for gender classification. The final classification of gender is done by summing the prediction of CNN on cropped faces.

### 3.1. Face Detection and Cropping Scheme

For both the training and testing image-set, we run the Cascade CNN face detector of Li et al. [20] to obtain the location of the face. After getting the location of the face, we increase the detected margin by 50% (50% of width left and right and 50% of height above.). The key factor behind increasing this margin and cropping scheme is the classification accuracy. Adding the margin and cropping helps in the classification accuracy.

For gender classification, the network needs the global view and alignment of a face. So for the increased marginal detected bounding box of the face, the cropping schemes of the face are shown in Figure 3. Detailed representation of our method is given below.

For any facial image, the image has the resolution of (i, j). After running the face detector of Li et al. [20], we get the detected bounding box coordinates (x1, y1) and (x2, y2). Then, increasing the margin of detected bounding box by 50%, we get the new coordinates (x1', y1') and (x2', y2'). Now, we divide the increased marginal detected bounding-box into two boxes. The two boxes are generated by adding and subtracting 100 to box coordinates. Thus Bounding box coordinates of the Left-box are (x1', y1') & (x2'-100, y2'-100). So, for the Bounding box coordinates of the Right-box are (x1'+100, y1'+100) and (x2', y2').

For three crop faces, we get the Left and Right crop as cropping the coordinates of Left and Right-box. For Middle crop, we take the intersection of Left and Right Box. We get the cropping coordinates of a Middle crop are (x1'+100, y1'+100) and (x2'-100, y2'-100).

For example, Adience image contains images with a resolution of 816 x 816. For one sample the increased marginal detected bound box has coordinates (211, 623) and (702, 310). Thus, Left-box has coordinates of (211, 623) & (602, 210) and Right-box has coordinates of (311, 723) & (702, 310). For the crops, Left and Right-crop are extracted using the box coordinates. As the Middle crop is the intersection of Left and Right-box, the cropping coordinates are (311, 723) and (602, 210). After getting the crops we squeezed crop-images to 224 x 224 pixels and used as input to VGG-16 architecture.

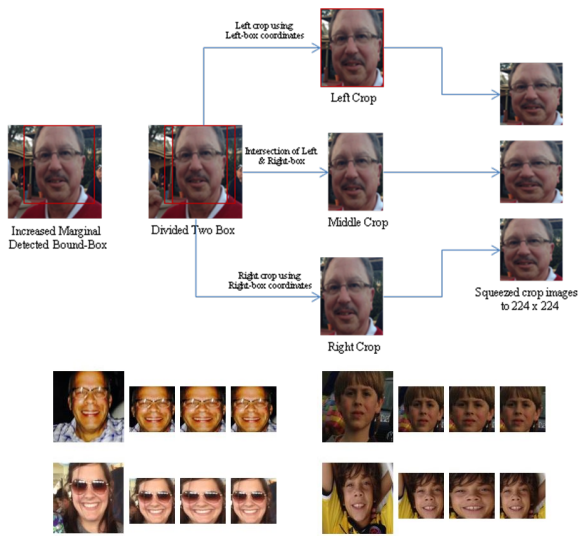

Figure 3: Cropping scheme for increased marginal detected bound-box on face, also sample images of cropped and squeezed faces are shown.

After getting the prediction from the network, we use the average summing technique to get the gender classification. For example in a sample image, the two crops show the result of Man, and one crop shows the result of a woman, we classify the image as Man. Opposite as the one crop shows the results of man, and two crops show the result of a woman, we classify the image as a woman.

### 3.2. Pre-processing

In general, taking two or more different dataset images usually results in a drop in performance due to the dataset bias problem [50]. In our case, the cropped images are variable, but the input dimension is usually fixed in a D-CNN. So for all cropped images, we squeezed and rescaled images to the dimension of 224 x 224 pixels.

### 3.3. Preparation of Training & Testing Data

We prepare the training, validation and testing image-set as per the given Table 2. We took care for splitting the data as the training and testing subject and images do not overlap. Notice that the LFW dataset has the resolution of 250x250. Thus, we scaled the LFW dataset to 816x816 resolutions, this is to prevent overfitting. After, using both datasets total 19906 images are for the training set, 5972 images are for validation set and 13935 images (9303 images of Adience and 4632 images of LFW) are for the testing set were split. So this makes the split of 50%, 15% and 35% for training, validation and testing image-set respectively. Table 2; shows the overview of training, validation, and testing image set.

### 3.4. Implementation Details

The implementation was carried on the workstation with Intel Core i7-6500 processor accelerated by NVIDIA Geforce GTX 1070.

VGG-16 [1] proposed by Simonyan et al., and it is 16-layer D-CNN used in the ILSVRC (ImageNet) [21] competition. The model achieves a 7.5% top-5 error rate on the validation set and placed second in the competition.

|  | Training | Validation | Testing |
|---|---|---|---|
| Adience | 13290 | 3987 | 9303 |
| LFW | 6616 | 1985 | 4632 |
| Total | 19906 | 5972 | 13935 |

Table 2: Overview of training, validation and testing image set split.

For implementation process, we initially define the fully connected layer and load the ImageNet pre-trained weight to the model. For the fine-tuning purpose, we truncate the original Softmax layer and replace it with our own. As gender has two classes (Man and Woman), we label the number of classes as two for gender classification task. We freeze the weight for the first 10 layers and so they remain intact throughout the fine-tuning process. It is common in computer vision to augment the training dataset and prepare many examples from original data. Some common augmentations are slight rotations, flipping images etc. The basic idea is to add small perturbations without damaging the central object so that D-CNN is more robust in real-world variations. We use the Keras [52] framework proposed by *Chollet et al.* with tensorflow as backend. As Keras provides the mechanism to perform such data augmentations quickly, thus we augment the training image-set with up to ± 5° of random rotation. We then fine-tune the model by minimizing the cross-entropy loss function as in [51] proposed by Bottou *et al.* A mini-batch size of 8 is used. We use Adam optimizer. During fine-tuning, we already have a pre-trained model which is very good, so we have used Adam optimizer with a very slow learning rate. We have observed little over-fitting during fine-tuning thus, we increased the dropout to 0.5. Notice that we use an initial learning rate of 0.001 rather than 0.01 (as it is normally used in the training from scratch). The model is fine-tuned for first 180 epochs with the learning rate of 0.001 and then another 20 epochs with the learning rate of 0.0001. The fine-tuning process takes more than half-day to complete.

## 4. Experiments

In this section, we present experimental evaluations of our proposed methods on the Adience and LFW datasets. We have done experiments on two folds, using our proposed method and without using cropping schemes. Our model is fine-tuned and tested on prepared data of Adience and LFW dataset as discussed in section 3.3.

### 4.1. Gender Classification on Adience and LFW

Initially, we have fine-tuned VGG-16 on prepared training data without using our proposed method, and check the performance on both the dataset, and as VGG-16 is pre-trained on ImageNet, we gained the classification accuracies of 83.2% and 89.2% on test set of Adience and LFW respectively, which is considerably good, but to further improve the classification accuracy, we have developed our method. After fine-tuning the network using the $2^B 3^C$ method, the classification accuracies increased by 7% (90.8%) and 6% (95.3%) on Adience and LFW respectively. Table 3: shows the gender classification results on Adience dataset. For overall on the training set accuracy 89.6% and on validation set accuracy is 86.3%. So 8447 out of 9303 images were predicted correct and 4414 out of 4632 images were predicted correctly for Adience and LFW test set respectively. Figure 4: shows the classification accuracy on prepared training and validation set of Adience and LFW dataset. Figure 5: shows the qualitative result on Adience and LFW test image-set. We haven't used the five-fold experiments, as we have found that for Adience the split is not good for the testing. As one of the fold contain 12256 training images, 1339 validation, and testing images. Thus to check the true ability of the network we have increased the number of validation and testing images to get the competitive results.

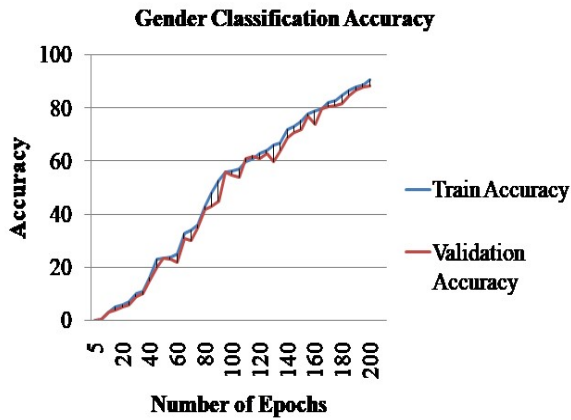

Figure 4: Gender classification accuracy on prepared training and validation image-set.

| Method | Accuracy |
|---|---|
| Best from [15] | 77.8% |
| Nine-patch [44] | 78.63% |
| Best from [46] | 79.3% |
| Single Crop [43] | 85.9% |
| Over-sample [43] | 86.8% |
| **Proposed using 2 Box 3 Crop** | **90.8%** |
| Best from [45] | 91% |

Table 3: Gender classification results on Adience dataset.

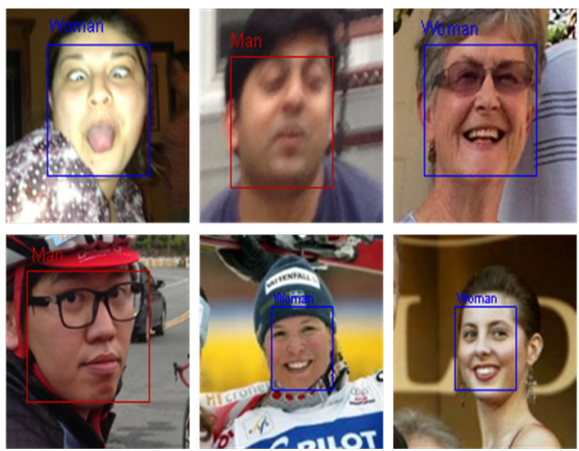

Figure 5: Qualitative results on some of the test image of Adience and LFW test image-set.

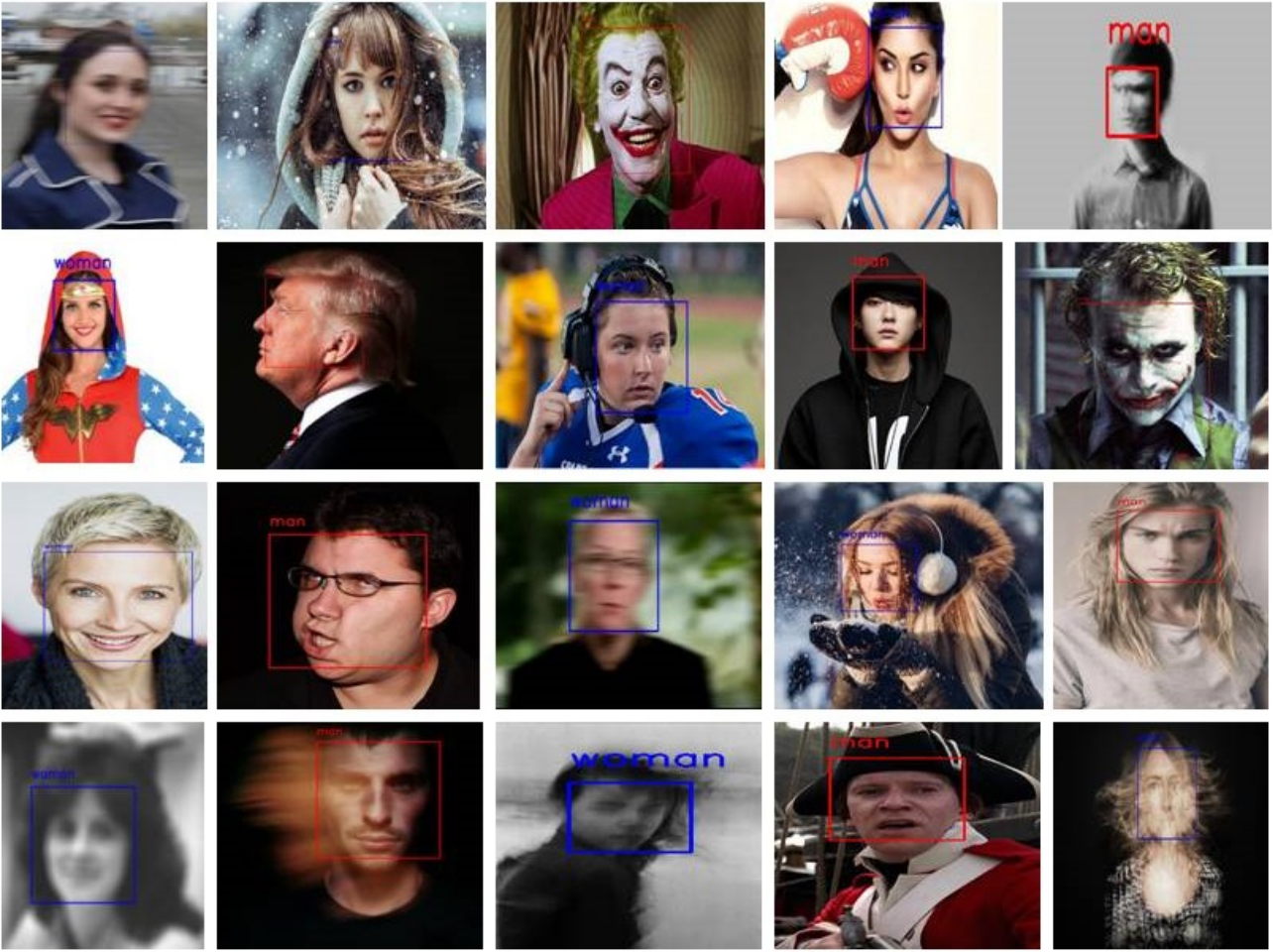

Figure 6: Qualitative results for the task of gender classification in different style of close-up images including noise, disguised variations and blur. All these tested images are never observed by network.
Red: Man Label, Blue: Woman Label.

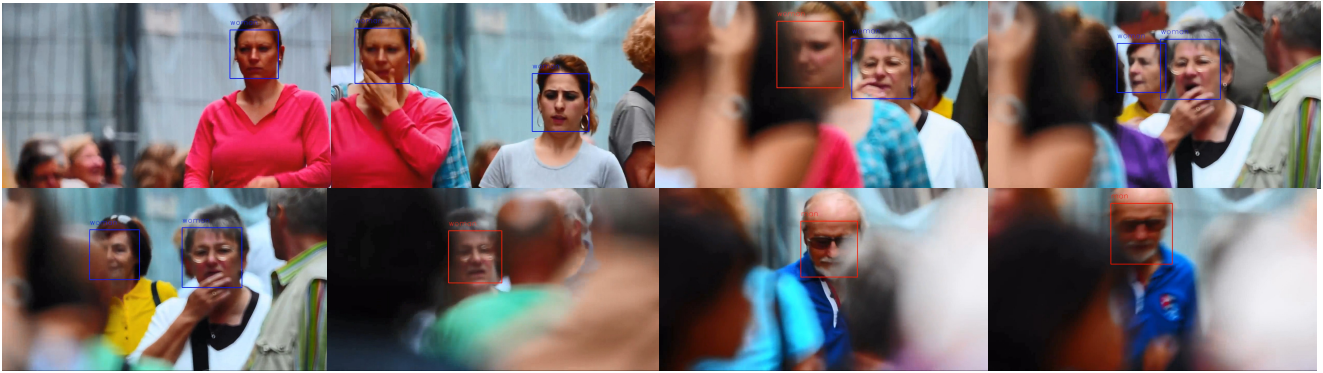

Figure 7: Examples of correct classification in surveillance video. Frame sequence is from Left to Right.
Red: Man Label, Blue: Woman Label.

To check the true ability of our network, we have done two more experiments. First, we have picked some of the test images in the different style of close-up. These test images are extremely wild, contains disguised variation on faces, blurred and extreme noise. Figure 6: shows the qualitative results of these tested images.

We have created a frame-based gender classification surveillance end-to-end system. In which, for every frame, the face-detector detect the faces, and as described for every faces our method applies and based on the prediction of fine-tuned CNN the final label will be predicted. Figure 7: shows the frame-based end-to-end classification example for one video sequence.

### 4.2. Limitations

We provide a few examples of gender misclassification in Figure 8. These show that many of the mistakes made by our network are due to very challenging viewpoints of some of the Adience and LFW test images. Mostly mistakes are due to a low-resolution image, blur, or heavy occlusions. Gender classification mistakes also occur for some images in which a human takes time to predict gender. Figure 8: shows some misclassified gender labels on test image of Adience and LFW.

One of the major drawbacks of our proposed method is that when we run the system on the surveillance video, the fps is too low with 7-10 fps. The reason behind this low fps is that in our method, after face-detector find the face with Bounding-box, the automatic box generation and cropping take too much computational power and also the prediction is based on average summing, thus it took time to classify final label.

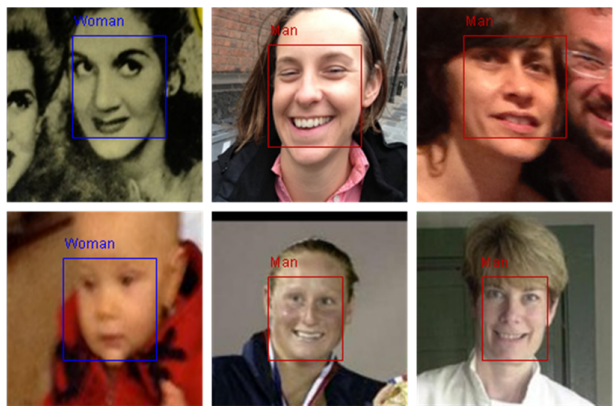

Figure 8: Gender misclassifications. Woman Subjects mistakenly classified as Man.

### 5. Conclusion

We tackled the classification of gender in still facial images and created an end-to-end frame-based system. Our method two boxes three crops ($2^B 3^C$) uses D-CNN with VGG-16 architecture pre-trained on ImageNet. To showcase the effectiveness of our method, we gain the gender classification accuracy of 90.8% and 95.3% on Adience and LFW dataset respectively.

Two major conclusions can be made from our experimental results. First, D-CNN can be used to provide improved Gender classification results considering the ablation study. Second, the simplicity of our method implies that more elaborate systems using more training data may well be capable of substantially improving results beyond reported here.